\documentclass[11pt]{article}

\usepackage[preprint]{acl}

\usepackage{times}
\usepackage{latexsym}

\usepackage[T1]{fontenc}
\usepackage[utf8]{inputenc}

\usepackage{microtype}

\usepackage{inconsolata}

\usepackage{graphicx}

\usepackage{makecell}

\title{Slovak Conceptual Dictionary}

\author{Miroslav Blšták \\
 Kempelen Institute of Intelligent Technologies \\
  \texttt{miroslav.blstak@kinit.sk}
  }

\begin{document}
\maketitle
\begin{abstract}
When solving tasks in the field of natural language processing, we sometimes need dictionary tools, such as lexicons, word form dictionaries or knowledge bases. However, the availability of dictionary data is insufficient in many languages, especially in the case of low resourced languages. In this article, we introduce a new conceptual dictionary for the Slovak language as the first linguistic tool of this kind. Since Slovak language is a language with limited linguistic resources and there are currently not available any machine-readable linguistic data sources with a sufficiently large volume of data, many tasks which require automated processing of Slovak text achieve weaker results compared to other languages and are almost impossible to solve.
\end{abstract}

\section{Introduction}
Machine-readable language resources greatly assist in automated text processing. Even in this era of large language models, dictionary approaches have their advantages and uses. They are more computationally efficient (less demanding on computing power and memory), so it can run less powerful devices. When using an indexed database, access to the item is handled in real time. Another advantage is the simple process of updating data: adding new words or error correction is much simpler in dictionary approaches as inserting or editing an entry in the dictionary is faster and no further modification is required (such as retraining in the case of language models). Last, but not least, transparency and explainability of results is more reliable when using deterministic approaches since we can exactly reveal how the algorithm works and how it calculates the result. So, at least for tasks where there is not a big difference in performance, using dictionary approaches seems more advantageous and is still a very important part of natural language processing tasks today.

A conceptual dictionary is a dictionary that groups items in taxonomy rather than arranging them in alphabetical order like typical lexicons. In many NLP tasks, it is not sufficient to work with textual data only on word level (text as a string) for various reasons. There are many words which are similar in meaning and different in word form (e.g. synonyms, hypernyms, diminutives etc.). Also, there are opposite situations, where one word form can express multiple meanings depending on the part-of-speech (POS) or context. We often need to disambiguate words before further text analysis. And also, there are many concepts which are represented by a sequence of words (known as multiword expressions, MWE), so a sequence of several words has a different meaning in comparison to using these words separately (e.g., “high school” or “hot dog” or “United States of America”). For these aforementioned situations and many others, it is necessary to process the whole text at semantic level. But first we need to extract the concepts from the text and identify their meaning (so-called mapping of words to concepts). Then we proceed to the next step and solve knowledge-based NLP tasks (e.g. sentiment analysis, text similarity, coreference resolution, named entity extraction or term extraction).

The main contribution of this paper is a dictionary of concepts for Slovak language. It is available through both a web interface and a machine-readable API. Currently, it is the largest dictionary of its kind for Slovak. It also contains additional tools, such as concept extractor from the text, visual concept map or word analogy generator. This ranks it among the unique projects of its kind for processing Slovak texts - for various tasks of extracting information from text as well as for tasks focused on semantic role labeling. During the design and development, we were partially inspired by popular dictionaries used in the field of NLP (e.g. Wordnet, ConceptNet, Wiktionary or DBPedia), and we also adapted the specific requirements of the language as rich word morphology and various types of ambiguities associated with this problem.

The rest of this paper is organized as follows. In the second part, we will summarize the current state of available dictionaries that can be used for semantic processing of Slovak text. In the third section, we will describe our proposed dictionary in detail. In the fourth part, we will present several examples of the use of this dictionary and its successful use in existing projects. In the last section, we will provide conclusions and our plans for the future work.

\section{Related Work}
There are currently no machine-readable dictionaries for the Slovak language that would contain a sufficient number of entries for use in text processing tasks. There are some dictionaries created by the Slovak Academy of Science\footnote{\url{https://slovnik.juls.savba.sk}}, but they have several drawbacks. Data are accessible only via web interface, so it is not possible to use within applications as they are not accessible via API. And it is not possible to download them and use them locally due to copyright restrictions. There is a small open-source community projects OpenThesaurus-SK\footnote{\url{https://www.sk-spell.sk.cx/thesaurus/}} created to support open source projects, but the number of items is very low and they have not been updated for the last few years. There is still a possibility to use foreign linguistic resources, and there are several attempts where dictionaries were created as an adaptation from multilingual projects (e.g. WordNet\footnote{\url{https://korpus.sk/korpusy-a-databazy/databazy/wordnet/}}~\citep{wordnet,skwordnet2011}, ConceptNet\footnote{\url{https://www.conceptnet.io/c/sk}}\footnote{\url{https://github.com/commonsense/conceptnet5/wiki/Languages}}~\citep{conceptnet}, BabelNet\footnote{\url{https://www.babelnet.org/search?word=&lang=SK}}~\citep{babelnet} or Wiktionary\footnote{\url{https://sk.wiktionary.org}}~\citep{meyer2012wiktionary}), but also the insufficient amount of data is very limited for use in real cases. Some potentially relevant multilingual projects are not available for Slovak language at all (e.g. DBpedia as its source data comes from Wikipedia which also doesn't have many terms in the Slovak version in comparison to other languages).

Also, when adopting language resources from other languages, there are several problems with handling specific features of the Slovak language (e.g. rich morphology and disambiguation). Disambiguation is particularly difficult in morphologically rich languages like Slovak due to the large number of forms for one lemma. Word in one form may have the same meaning as another word in a different form, and collisions also occur across different parts of speech or genders (so it is appropriate to store not only the basic form of the word and its other forms, but in some cases the gender of the word is also necessary for disambiguation and determining the meaning).

In table~\ref{comparisons}, we compare the most beneficial language resources containing Slovak entries which are available to use. We excluded those tools that include copyright restrictions and it is impossible to use them not even for research purposes.

\begin{table*}
  \centering
  \begin{tabular}{lcccc}
    \hline
    \textbf{Dictionary} & \textbf{ConceptNet} & \textbf{Wiktionary} & \textbf{OpenThesaurus-SK} & \textbf{SK-WordNet} \\
    \hline
    Slovak entries &30~000&37 700&13 500&25 000\\
    POS 
    &\makecell{nouns,\\adjectives,\\verbs\\} 
    &\makecell{nouns,\\adjectives,\\pronouns,\\numerals,\\ verbs}
    &\makecell{nouns,\\adjectives,\\verbs}
    &\makecell{nouns,\\adjectives,\\adverbs,\\verbs}\\
    Date of creation & 1999&2004&2004&2013 \\
    Last update     & 2021&2025&2008&2013 \\
    Disambiguation &by POS&by POS& no disambiguation& via English translation \\
    \hline
  \end{tabular}
  \caption{\label{comparisons}Comparison of existing dictionaries for the Slovak language. 
  }
\end{table*}

However, neither tool has solved the issue of ambiguity of terms, where one word can refer to different concepts (has multiple meanings). Some tools provide information about the part of speech, but this is not enough to distinguish between multiple meanings. There are many words that have multiple meanings with the same POS tag. For example, ConceptNet distinguishes between terms `train` if the part of speech tag is different (train as a noun\footnote{\url{https://conceptnet.io/c/en/train/n}} and train as a verb\footnote{\url{https://conceptnet.io/c/en/train/v}}) but when the word has several meanings with the same part of speech tag, all meanings are merged together (e.g. “run” in context of moving: “run a marathon” and in context of starting some project “run a business”\footnote{\url{https://conceptnet.io/c/en/run}}). Word sense disambiguation is still a difficult NLP problem in both cases - even within the same POS tag \citep{Abraham2024, Bevilacqua2021}.

Based on an analysis of the current state of dictionary data usable for Slovak text processing, we identified several weaknesses and shortcomings:

\begin{itemize}
  \item There is currently no dictionary of concepts created for the Slovak language that would contain a sufficient number of entries.
  \item Current dictionaries do not have a machine readable interface, so they are hardly used for automated text processing. Some of them are available through a web interface, but data cannot be downloaded and used due to licensing restrictions.
  \item The use of dictionaries from other languages is problematic. Existing multilingual tools do not take into account the specific features of Slovak (especially complicated morphology and ambiguity without specifying part of speech or grammatical categories), so the use of automatic translation or mapping words to other language reduces the accuracy of processing.
  \item Based on previous statements, we currently do not have the ability to solve text processing tasks in Slovak that require a dictionary of concepts - neither for research nor for commercial use.
\end{itemize}

\section{Our Proposed Dictionary of Concepts}
\begin{figure*}[t]
\centering
  \includegraphics[width=120mm]{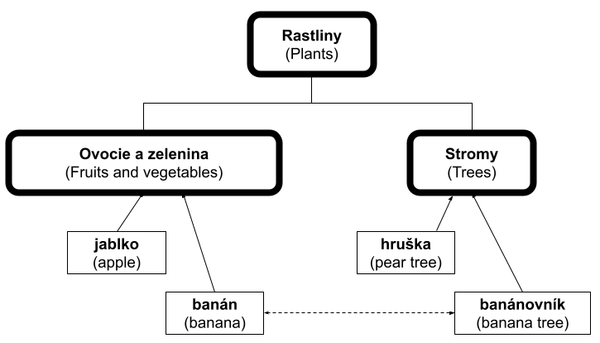}   \caption {example of data structure (concepts in taxonomy structure and relationships between concepts)}
  \label{fig:img1}
\end{figure*}

In this chapter, we introduce the structure and data of our conceptual dictionary. The basic unit of the dictionary is a concept (or term) arranged in a hierarchical structure of categories with links to other concepts by named relationships. There are three main entities in our data model:
\begin{itemize}
  \item concepts (represented by lemma and namespace),
  \item categories in hierarchical structure (taxonomy),
  \item relationships between the couple of concepts.
\end{itemize}
The main entity of the data model is a concept represented in the form of a lemma (as a basic form of the word). Each concept is placed in a hierarchical structure of categories (taxonomy). As the categories are in a hierarchical structure, each concept belongs to one or more categories (as it is shown in Figure~\ref{fig:img1}). The actual hierarchy of categories is shown in documentation\footnote{\url{https://pojmy.kinit.sk/stats/namespaces}}. The primary key of a dictionary entry is a combination of concept (in lemma form) and its full path in taxonomy separated by slash character. This is due to the fact that many different concepts with the same form of lemma should have multiple meanings (e.g. Slovak word “hruška” means pear as fruit and also pear as tree). The concept in Slovak language is sometimes represented as a multiword expression (MWE), which means that it consists of a sequence of words (e.g. high school). In this case, the concept in the dictionary is stored in a readable lemmatized form. It means that concepts which consist of a pair of words where the noun is preceded by an adjective, the form of the adjective word has the same form as the noun. In the Slovak language, the form of the adjective should contain many different variants (adding affix to stem of the word) due to word inflection, but using the lemmatized form of adjective in MWE is sometimes hardly readable. So, the noun is in lemma form and the adjective inherits its form due to readability. This type of problem does not exist in languages like English where there are only two forms for a noun (singular and plural). 

There are also additional information for concepts based on POS tag: 
\begin{itemize}
  \item alternative forms of lemma (linked to external morphological dictionary): as the Slovak language is morphologically rich language, it helps a lot to know all possible forms of concept,
  \item translations to foreign languages (translated automatically using a translation service from Google into 11 most common languages used within Slovak texts: English, Czech, German, Hungarian, Polish, Croatian, French, Italian, Spanish, Portuguese, and Latin),
  \item Part-of-Speech tag (POS tag): which may helps to distinguish between two possible meanings of a word (like the English word “fly” can represent a type of insect or flying activity). Currently we store mainly nouns, adjectives, numerals, verbs and adverbs, 
  \item gender attribute: which may helps to disambiguate the meaning of a word (for example the same word should have different meanings in different gender - e.g., word “zver”).
\end{itemize}
And there are also two special types of entries (except one word entries and MWE): 
\begin{itemize}
  \item characters (e.g., asterisk, mathematical signs etc.): linked to their word equivalent (e.g., concept “plus” joined to character “+”),
  \item emoticons: concepts are occasionally expressed by a graphic symbol instead of a word (e.g., pear as emoji).
\end{itemize}
Relationships between the couple of concepts are the second important part of the dictionary. Currently, we store more than 20 basic relationships between concepts which are common in similar tools (e.g., synonyms, antonyms, hypernyms, diminutives, …). The complete list of relationships can be found on the project website\footnote{\url{https://pojmy.kinit.sk/stats/relationships}}. Some of the relationships are symmetric (e.g. synonyms or antonyms) and some are one-way resp. they have a different meaning when reading from the opposite side (e.g. diminutive or augmentative).

Although there are essentially no freely available dictionaries for the Slovak language, fortunately, we did not have to start from scratch. We used dictionary data from the creators of the Slovak educational project Eduself.sk\footnote{\url{https://www.eduself.sk/}}, who provided them to us for this project. They are the authors of several dictionary data that they use in their educational applications and they provided us with their taxonomic dictionary, which we used as the basis for building our concept dictionary. Then, we expanded the list of concepts and added relationships - partly manually and partly generated with hand-written scripts. Concepts and relationships generated by scripts were also validated by humans. Only automated translations of concepts into foreign languages have not yet been checked, we may use them in the future to connect with other word databases - but we are not working on this yet. Dictionary data were improved over time as we used the dictionary for various NLP tasks. The first version was created in the summer 2021. After several years of development and testing, we are publishing it. The dictionary is a joint work of our institute (Kempelen Institute of Intelligent Technologies) and the authors of the Eduself.sk project. We have agreed to make the dictionary available via web interface and also via machine-readable API.

\section{Application}
 A dictionary of concepts can be used to solve various types of tasks, either alone or in combination with alternative techniques. A straightforward use is in tasks focused on extracting information from a text: extraction of named entities (names of people, places, countries) or for extracting general concepts from a domain (extraction of foods, drugs, diseases, animals, etc.). Also in identifying relationships between different words in a text (e.g. coreference resolution), when using different forms of the same name (for example, the names Miro and Mirko refer to the same name in Slovak, although the form of the word is slightly different). Information about relationships between words (especially synonyms or alternative forms of the same word) can help in calculating the semantic similarity of two texts. Last but not least, we can use stored relationships between concepts to generate datasets for model validation and bias detection - for example, by generating pairs of sentences and observing how an existing model changes its prediction when changing a person's first name, date of birth, gender or some other type of word.

\begin{figure*}[t]
  \includegraphics[width=160mm]{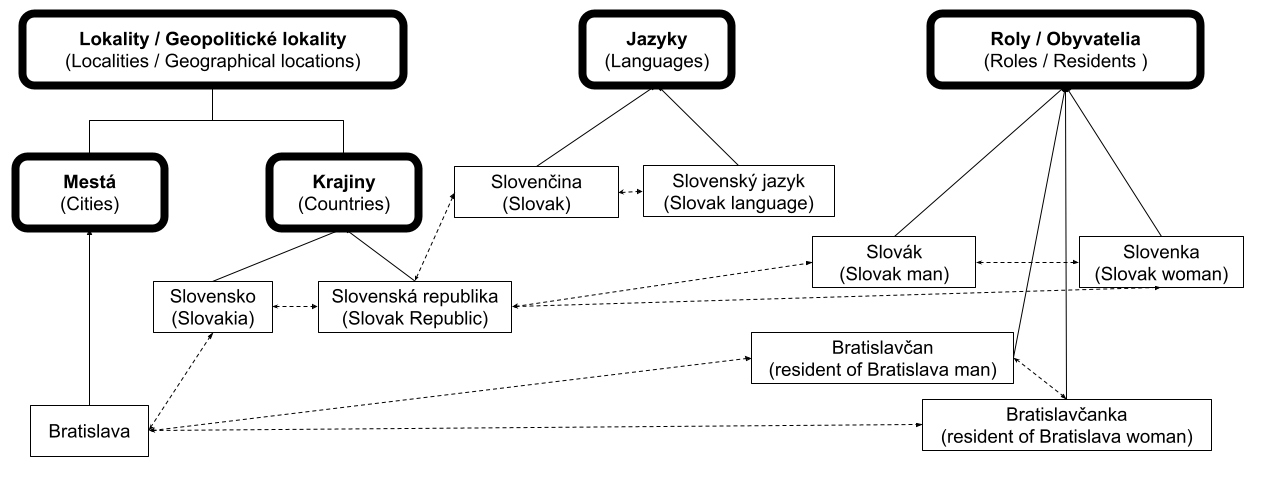}   \caption {Example of a conceptual network.}
  \label{fig:img2}
\end{figure*}

We already used our dictionary in several internal projects, and also in one research project focused on the topic of migration. We wanted to extract location information from media articles which are somehow related to migration paths and specific information about migrating people (e.g., nationality, race, religion) \citep{blstak-etal-2025-dictionary}. We have used this dictionary for solving two tasks: detect the source and destination country of migration and to find out if the migration is somehow related to Slovakia or Slovaks (not only if the migration source or destination is Slovakia, but also if the migration influences the public opinion of Slovak readers). These tasks are similar to typical named entity recognition tasks (NER), however there are several important differences which cause that typical NER algorithms are not enough. References to a country can be mentioned in the text not only directly (mentioning the name of the country), but much more often indirectly through a municipality, district, or by crossing a border (for example, a river, mountain range, etc.). Also extracting a resident of a certain area is not trivial in languages like Slovak, since these concepts are created using various suffixes different for male and female gender, so words representing relations to concepts are usually not identified by a common NER tool. Another common way to mention locations is through adjectives with no capital letter in Slovak language, so it is difficult to  (e.g. the concept “mayor of Bratislava” is mentioned as  “bratislavský primátor”). Our experiments have shown that by using relationships from the conceptual dictionary, we can extract this information more accurately than conventional methods and also more accurately than by large language models. Results of our experiments confirm the advantage of using these approaches for information extraction tasks. In the figure~\ref{fig:img2}, there is an example of a semantic network for concepts related to location.

The usage of linguistic tools is not limited only to linguistics and natural language processing tasks. Another significant area where these types of tools are lacking for the Slovak language is e-learning and computer-assisted learning (CAL). For example, preparing for standardized tests or creating educational quiz questions are popular tasks where linguistic resources can be used. Word analogies (known also as SAT analogies) are used for college admissions or verbal intelligence tests. Proportional analogies are statements of the form: a1 is to b1 as a2 is to b2 (e.g.: man $\rightarrow$ king $\iff$ woman $\rightarrow$ queen). 
The task of filling in missing values of a proportional analogy is common parts of various verbal tests.

\section{Conclusion and Future Work}
In this article, we have introduced our conceptual dictionary for Slovak language, which can help to solve various text processing tasks. It is available at \url{https://pojmy.kinit.sk}. The structure of data in the dictionary and its schema are inspired by existing linguistics tools, but they are adapted for the specific problems of Slovak as a language with rich morphology. The dictionary entries (concepts) are arranged in a hierarchical structure of categories, which also helps to disambiguate terms that have multiple meanings. Currently, there are more than 140 000 concepts and more than 300 000 relationships between these concepts stored in our database, and we are regularly inserting new data. In terms of the volume of entries, our dictionary is already the largest compared to other tools for Slovak which are available. Storing information about Part of Speech and gender (within some part of speech categories) is necessary in the case of Slovak due to the diversity of forms and situations where the same term takes on multiple meanings with respect to grammatical categories. We also keep alternative forms for most words for at least two reasons: in ordinary Slovak text, words are usually found in forms other than the basic form, and it also often happens that the same word (the same sequence of letters) has different meanings in different forms - so each form refers to a different concept. The dictionary also includes multi-words expressions, since a concept consisting of multiple words should have different meanings in comparison to the meaning of these words mentioned separately. And finally, the concept database also includes emoticons and symbols, as they are widely used in informal text, where they are an important part of the text for semantic analysis. Relationships between tuples of concepts play an important role in semantic analysis as they reveal hidden meanings behind the words. 

We believe that this new linguistic tool can significantly help in tasks focused on extracting information and knowledge from text. There is also the first version of API available, and we plan to extend its functionality in the near future. Currently, there are no other tools with a similar scope available for the Slovak language.

In the future, we plan to extend the API, adding more data (concepts and relationships) and develop tools based on dictionary data (e.g., rule-based dataset generation, number to text transformer and similar text transformation tasks).

%\section*{Acknowledgments}
%This document has been adapted

\bibliography{anthology}

\appendix
\label{sec:appendix}

\section{Appendix: Basic Overview of Data}
This appendix presents the dictionary's statistical data (December 2025). Approximate (rounded) number of data in our dictionary is shown in the table~\ref{table1}.
\begin{table}[h]
\centering
\begin{tabular}{lrl}
    \hline 
    number of concepts &145 000 \\
    number of categories &166 \\
    relationships between concepts &355 000 \\
    \hline 
\end{tabular}
\caption{Current number of entries in the dictionary database}
\label{table1}
\end{table}
The current list of categories\footnote{\url{https://pojmy.kinit.sk/stats/namespaces}} and the list of relationships\footnote{\url{https://pojmy.kinit.sk/stats/relationships}} is available on the dictionary website, including other statistics on the current number of entries. Approximate (rounded) number of concepts grouped by tag is shown in the table~\ref{table2}.

\begin{table}[h]
\begin{tabular}{lrl}
    \hline 
    concept type & quantity & examples (translated into English) \\
    \hline 
    Nouns & 69 000 & pes (dog), strom (tree), noha (leg)\\
    Adjectives & 50 000 & silný (strong), čítajúci (reading) \\
    Numerals & 1 500 & sedem (seven), trinásť (thirteen) \\
    Verbs & 7 700 & čítať (to read), písať (to write) \\
    Adverbs & 150 & dobre (well), blažene (blissfully) \\
    MWE & 2 200 & vysoká škola (high school), New York\\
    \hline 
\end{tabular}
\caption{Current number of concepts grouped by POS tag.}
\label{table2}
\end{table}

\end{document}